\documentclass{article}
\usepackage{spconf,amsmath,graphicx, svg}

\usepackage{adjustbox}
\title{MMSR: MULTIPLE-MODEL LEARNED IMAGE SUPER-RESOLUTION \\
BENEFITING FROM CLASS-SPECIFIC IMAGE PRIORS}
\name{Cansu Korkmaz\sthanks{This work was supported in part by an AI Fellowship to C. Korkmaz provided by the KUIS AI Center.}, A.Murat Tekalp\sthanks{This work was supported by in part by TUBITAK 2247-A Award No.~120C156 and a grant from Turkish Is Bank to KUIS AILab. AMT also acknowledges support from Turkish Academy of Sciences (TUBA).}, Zafer Do\u{g}an\sthanks{Z.D. acknowledges that this work was supported in part by TUBITAK 2232 International Fellowship for Outstanding Researchers Award (No. 118C337) and an AI Fellowship provided by the KUIS AI Lab.}}
\address{Koç University, Dept. of Electrical and Electronics Engineering, Istanbul, Turkey \\
\{ckorkmaz14, mtekalp, zdogan\}@ku.edu.tr}
\begin{document}
%
\maketitle
\begin{abstract}
Assuming a known degradation model, the performance of a learned image super-resolution (SR) model depends on how well the variety of image characteristics within the training set matches those in the test set. As a result, the performance of an SR model varies noticeably from image to image over a test set depending on whether characteristics of specific images are similar to those in the training set or not. Hence, in general, a single SR model cannot generalize well enough for all types of image content. In this work, we show that training multiple SR models for different classes of images (e.g., for text, texture, etc.) to exploit class-specific image priors and employing a post-processing network that learns how to best fuse the outputs produced by these multiple SR models surpasses the performance of state-of-the-art generic SR models. Experimental results clearly demonstrate that the proposed multiple-model SR (MMSR) approach significantly outperforms a single pre-trained state-of-the-art SR model both quantitatively and visually. It even exceeds the~performance of the best single class-specific SR model trained on similar text or texture images. 
\end{abstract}


\section{Introduction}
\label{sec:intro}
\vspace{-3pt}
Image super-resolution (SR) aims to generate the latent high-resolution (HR) images from low-resolution (LR) ones by generating information that does not exist in the LR images~\cite{10.5555/2843012,kim2016accurate}. 
Single-image SR (SISR) by learning from examples dates back to the seminal works of Freeman \cite{Freeman2000, Freeman2002}, which implicitly learned features from a database of HR and LR image patch pairs. These example natural image patch pairs were employed as data-driven priors to extrapolate high frequency image details. Example-based SISR has later been applied to target specific classes of images, e.g., face images \cite{Kanade2000, Freeman2007}. These class-based SISR algorithms have yielded superior results, because they were able to better capture class-specific image priors. 

\begin{figure}[t!]
\begin{minipage}[b]{0.72\linewidth}
  \centering
  \centerline{\includegraphics[width=6.1cm]{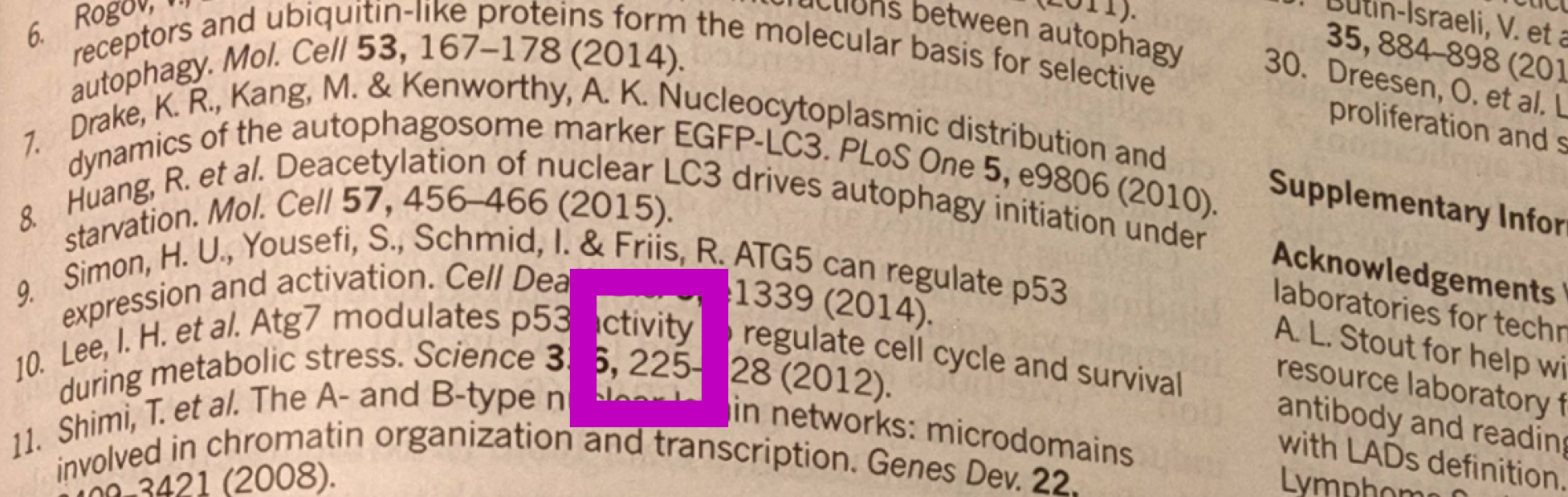}}
  \centerline{(a)}\medskip
\end{minipage}
\begin{minipage}[b]{.24\linewidth}
  \centering
  \centerline{\includegraphics[width=2.0cm]{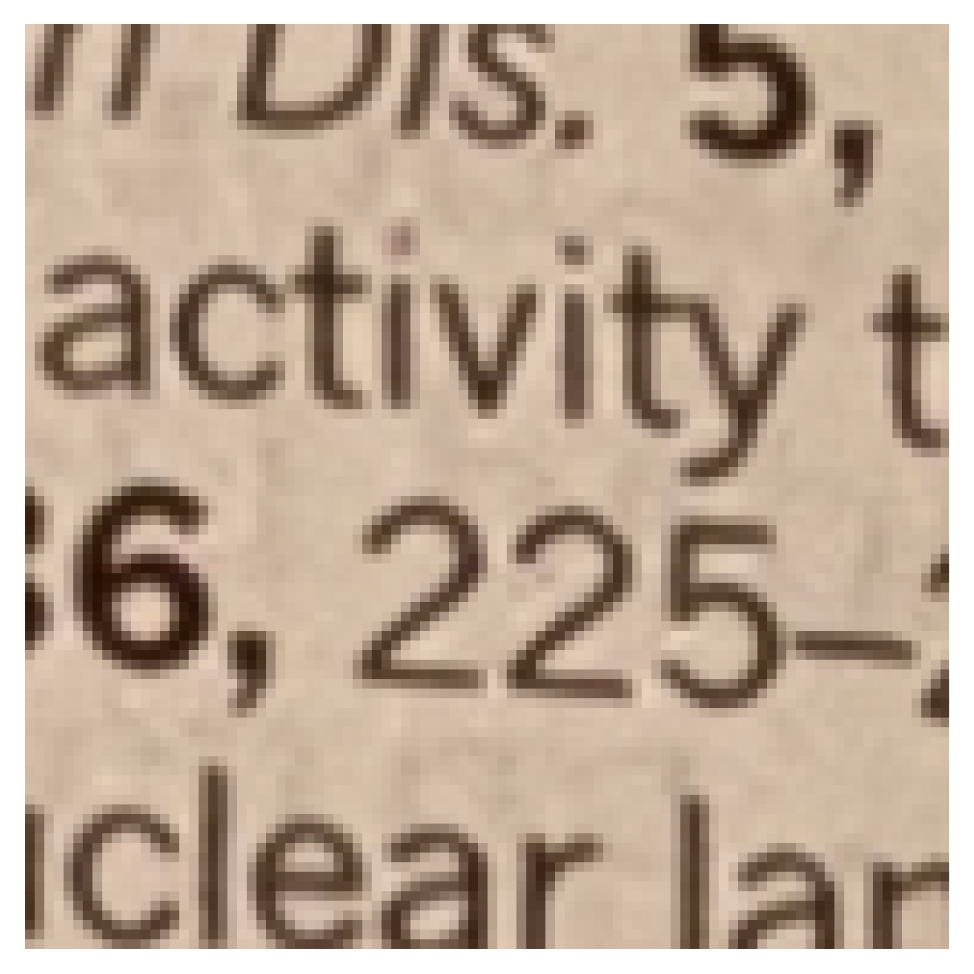}}
  \centerline{(b)}\medskip
\end{minipage}
\begin{minipage}[b]{.24\linewidth}
  \centering
  \centerline{\includegraphics[width=2.0cm]{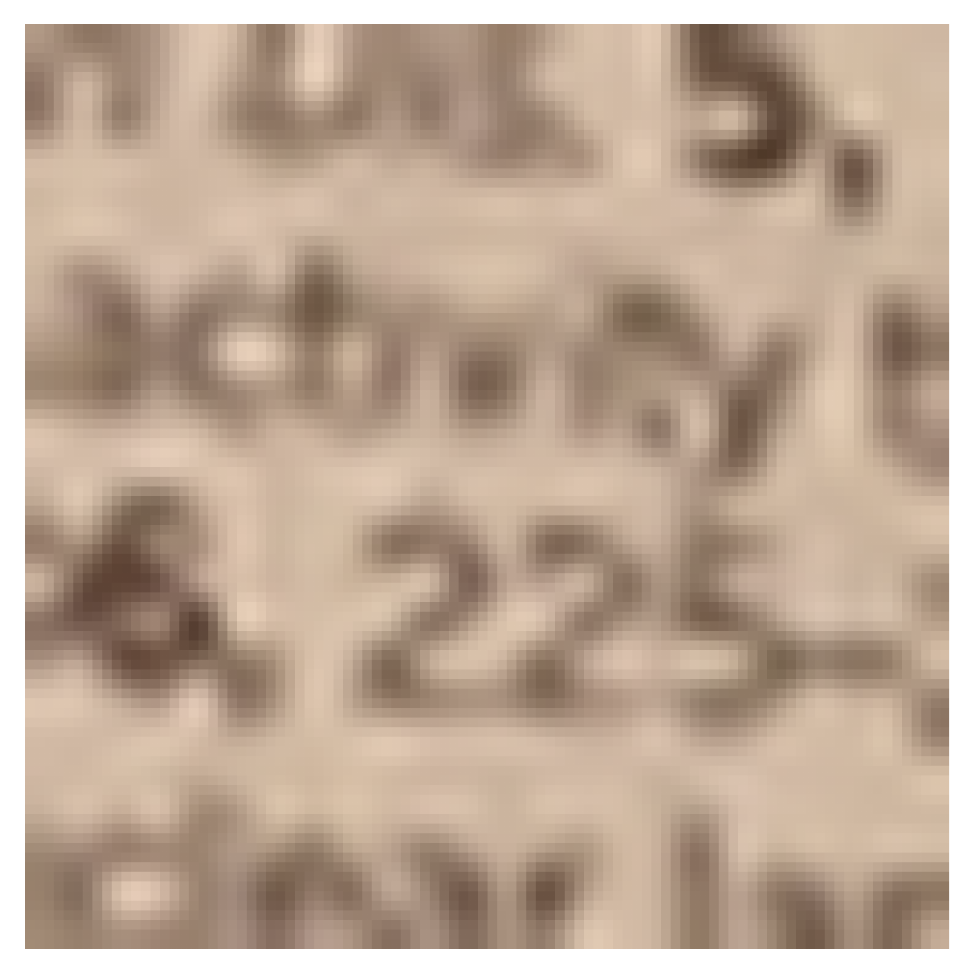}}
  \centerline{(c)}\medskip
\end{minipage}
\hfill
\begin{minipage}[b]{0.24\linewidth}
  \centering
  \centerline{\includegraphics[width=2.0cm]{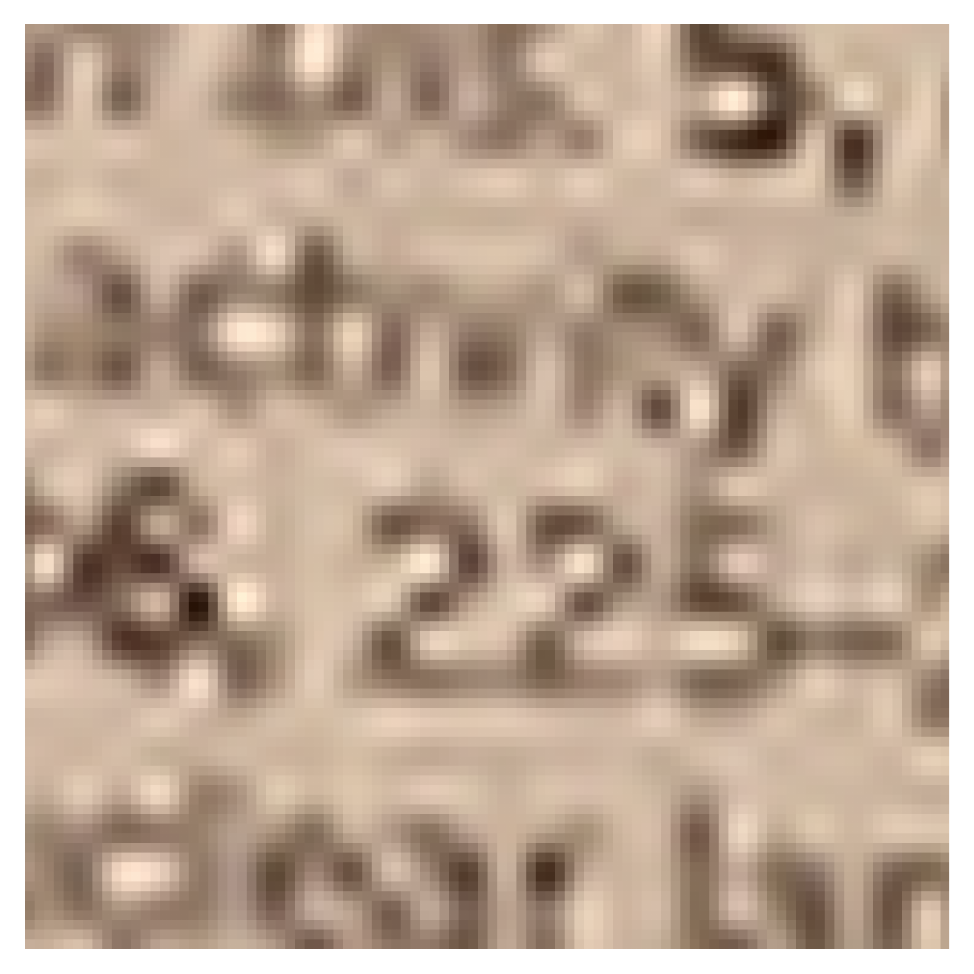}}
  \centerline{(d)}\medskip
\end{minipage}
\begin{minipage}[b]{.24\linewidth}
  \centering
  \centerline{\includegraphics[width=2.0cm]{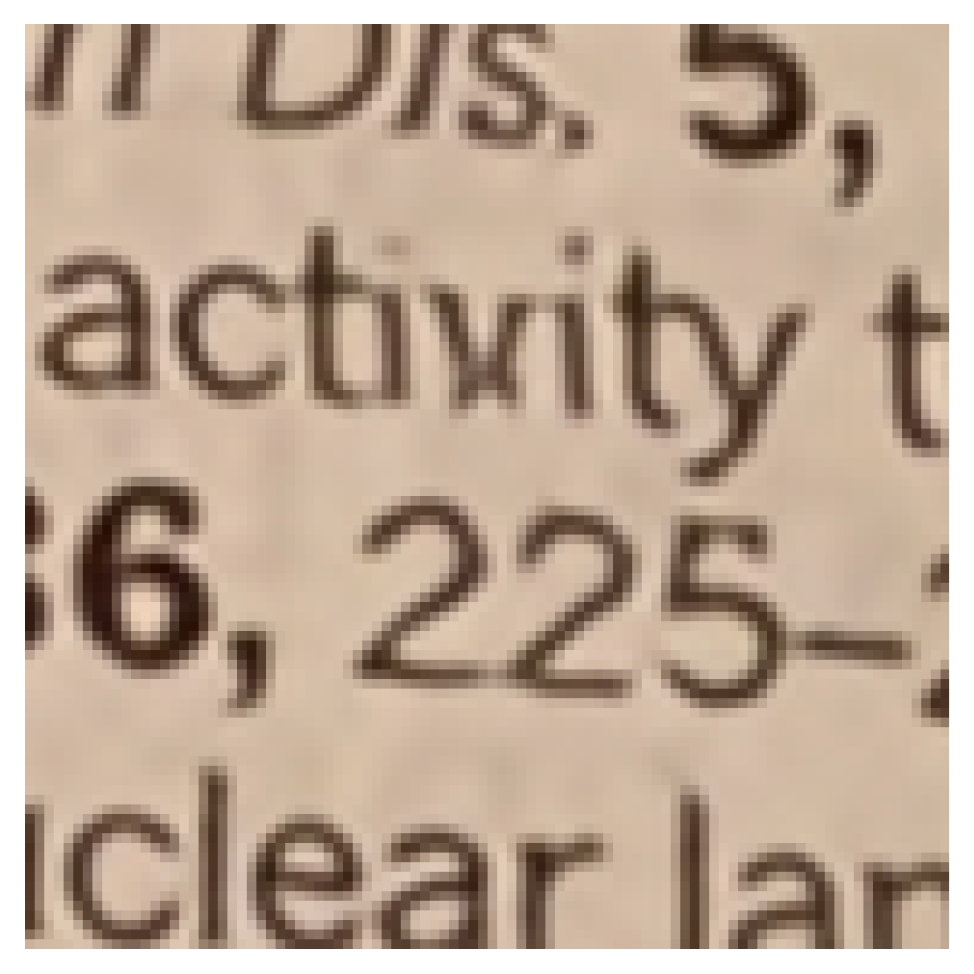}}
  \centerline{(e)}\medskip
\end{minipage}
\hfill
\begin{minipage}[b]{0.24\linewidth}
  \centering
  \centerline{\includegraphics[width=2.0cm]{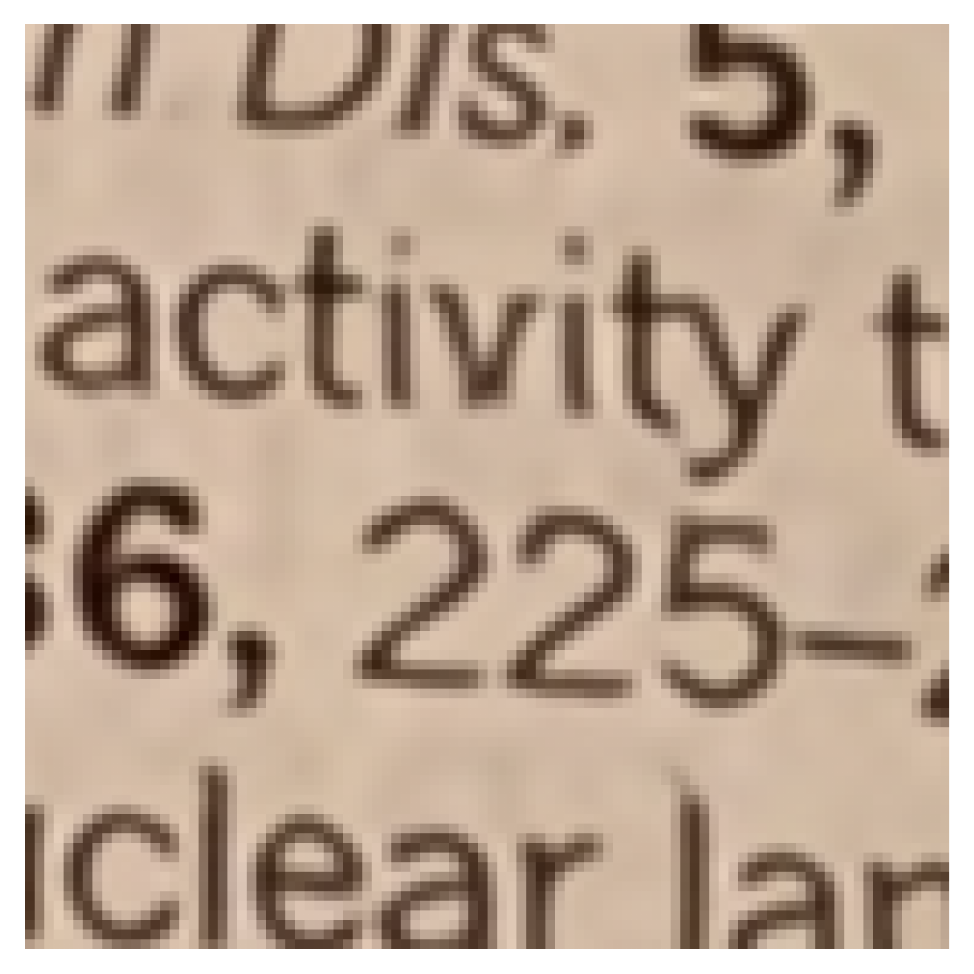}}
  \centerline{(f)}\medskip
\end{minipage}  \vspace{-15pt}
\caption{This example demonstrates that $\times$4 SR by class-specific training for text images, not only yields higher PSNR, but also better visual quality. (a) HR image ImagePairs \cite{joze2020imagepairs} test set, (b) cropped HR patch, (c) bicubic interpolation of $\downarrow$4 image, (d) ZSSR - 23.74 dB, (e) generic EDSR model - 31.86~dB, (f) text-class EDSR model - 32.574 dB}
\label{fig:results}
\end{figure}

Existing learned SISR methods \cite{dong2015image,kim2016accurate, ledig2017photorealistic, ZSSR, EDSR2017} have achieved remarkable results by training a generic SR model using a set of HR and LR image pairs. However, intensity patterns in different types of images vary significantly. For instance, a text image contains sharp edges against a uniform background, while an outdoor image contains fine textures, such as trees, grass, etc. As a result, capturing image priors by means of a single generic SR model does not yield the best results for a variety of classes of images. To summarize, drawbacks of training a generic SR model for all types of images are: (i) different images have different characteristics,  (ii) even in a single image, different regions (patches) usually have distinct features, such as text and texture within the same image. Hence, a generic SR model is not adequate to learn different types of image priors and the one-model-for-all approach is sub-optimal, leading to a large performance variance over the test dataset.


\noindent \textbf{Contributions:} Main contributions of this work are: \textbf{i)} We propose class-specific SR models trained for specific classes of images, which have similar characteristics, to better capture image priors and decrease performance variance across the test dataset. \textbf{ii)} To address the variation of image characteristics within a single image, we propose a multiple model SR (MMSR) framework with a post-SR fusion network that learns how to combine SR images obtained by multiple SR models, each tuned to a specific class of image priors.


\section{Related Work}
\label{related}
\vspace{-3pt}

Many researchers, starting with early work on SRCNN \cite{dong2015image} up to more recent advanced architectures such as EDSR \cite{EDSR2017}, RCAN \cite{RCAN2018} and others \cite{kim2016accurate, ledig2017photorealistic, Sun_2020}, have demonstrated superior performance of supervised deep learning for the SISR task. 
A~limitation of these approaches is that they all rely on a single generic SR model for all images in the test set. However, a model trained on a particular training set cannot generalize well for images with different characteristics even in the case of a known degradation model. Accordingly, it was observed in \cite{8552961} that the mean PSNR or SSIM values reported over a test set using a single SR model display a large variance, where the PSNR values may vary up to 5 dB for different images in the test set. This observation and the success of early class-based image hallucination methods~\cite{Kanade2000,Freeman2007} motivate us to explore SR models that exploit class-specific image priors.

There are also unsupervised deep learning methods, such as KernelGAN \cite{Kernel} and Zero-Shot Super-Resolution Using Deep Internal Learning (ZSSR) \cite{ZSSR}, which provide successful results for single image SR. In the extreme, one can consider designing a different inference model for each individual image as in the case of unsupervised deep learning methods. Specifically, ZSSR generates an image-specific model taking advantage of self-similarity and internal statistics of images and does not rely on an external training set. However, zero-shot models need to be trained for each test image individually and their performance is in general lower compared to those obtained by supervised learning.



\section{MMSR: Multiple-Model SR Network}
\label{sec:method}
\vspace{-3pt}

 
 
\begin{figure}[t]
  \centering
  \includegraphics[width=0.35\textwidth]{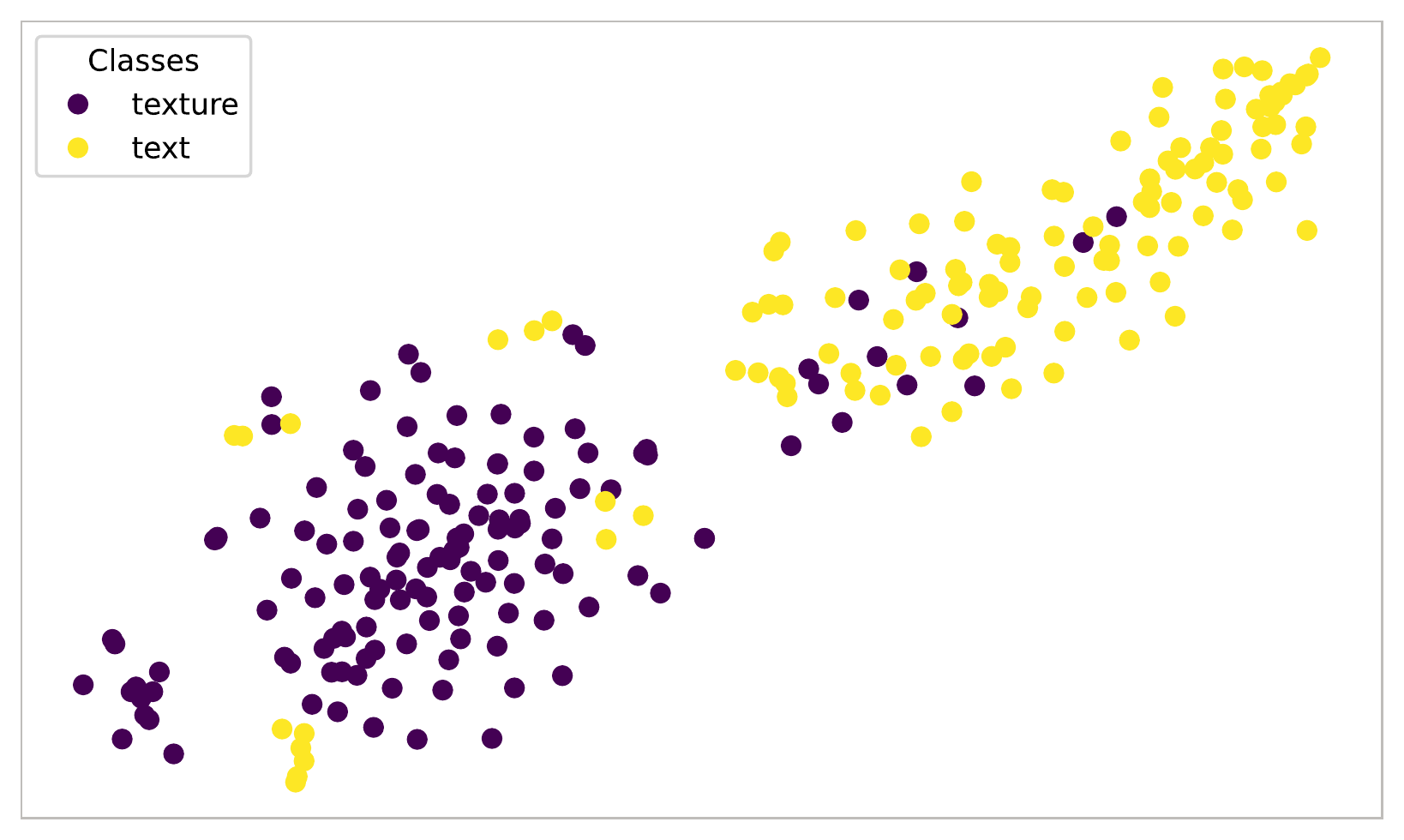}
\vspace{-10pt}
\caption{Visualization of the training data consisting of a mixture of natural texture patches from DIV2K and text patches from ImagePairs~\cite{joze2020imagepairs} using t-distributed stochastic neighbor embedding (t-SNE)~\cite{Maaten2008VisualizingDU} shows approximately two clusters.}
\label{fig:clustering} \vspace{-6pt}
\end{figure}

The standard paradigm for training an SR network is to learn a single generic model to capture both the degradation and the image prior. Assuming the degradation model is the same for all images in the training and test sets, we investigate whether a single model can satisfactorily capture image priors for a diverse set of images. The 2D t-SNE \cite{Maaten2008VisualizingDU} plot for patches from our training data, depicted in Fig.~\ref{fig:clustering}, shows at least two well-separated clusters, suggesting that a single generic model may be suboptimal for accurately capturing image priors for all images in this training set. Hence, we propose multiple SR models each tuned to class-specific image priors. To better exploit this concept, in this paper, we propose a multi-model SR network architecture including a post-processing network that learns how to fuse the output SR images produced by each model. Interestingly, our results indicate that even in the case of homogeneous input images fusing the results produced by these multiple class-specific models by a post-processing network outperforms the performance of the best class-specific model trained for the particular class of images. 

\begin{figure}[b!]
\hspace{-8pt}  \includegraphics[width=0.5\textwidth]{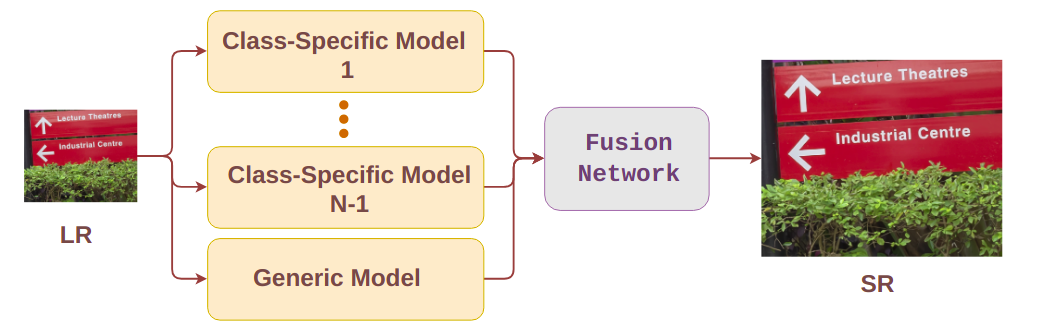}
\vspace{-15pt}
 \centering 
 \caption{The proposed multiple-model SR (MMSR) network.}
\label{fig:fusion_arch}
\end{figure}

The proposed MMSR architecture, depicted in Figure~\ref{fig:fusion_arch}, consists of a bank of $N$ SR models, where $N-1$ of them are class-specific SR models, each tuned to image priors from a specific class of images, and the last one is a generic SR model (representing the class ``other"), as well as a post-SR network that learns how to best fuse the outputs of these multiple SR models. Our results show that the proposed MMSR framework is not only effective in super-resolving images with mixed content, e.g., text and texture in the same image, but also yields results better than a generic or class-specific model on a single class image, such as a text-only image.

\vspace{-12pt}
\subsection{Class-Specific and Generic SR Models}
\label{sec:class_generic}
\vspace{-5pt}
We employ standard SR architectures, such as EDSR~\cite{EDSR2017} and RCAN~\cite{RCAN2018}, for both class-specific and generic SR models. In our experiments, due to its simplicity, each class-specific SR model and the generic SR model (representing the class ``other") has been an instance of the EDSR baseline network with 16 residual blocks and feature size 64 channels. For the $\times $4 SR task, we employ 2 successive $\times $2 pixel-shuffler layers~\cite{shi2016realtime}. The number $N$ of class-specific SR models can be determined according to the application.

\subsection{Class-Specific Fusion} 
\label{sec:res_mmsr}
When the target test set contains a specific class of homogeneous images, e.g., text only or texture only, then the fusion model can be trained by using patch pairs representing only that class. More specifically, first each SR model is trained on a class-specific training set to benefit from class-specific image priors and then a fusion model is trained on a particular class-specific training set, e.g., text-only, using output images from both class-specific SR models with text images as input. We refer to this case as class-specific fusion training. Our results show that the proposed MMSR with class-specific fusion yields images superior to that can be obtained by the best single model trained on a class-specific training set. This reveals even homogeneous images can benefit from MMSR. 
\vspace{-8pt}
\subsection{Generic Fusion with Pre-Trained SR Networks} When the target test set contains non-homogeneous images, which contains multiple classes e.g., text and texture patches within the same image, we propose training the fusion module using a mixture of HR and LR patch pairs from multiple class-specific training sets. More specifically, we again have $N$ class-specific SR models trained on respective class-specific training sets. But this time, different from class-specific fusion, the fusion model is trained by using a mixture of patch pairs coming from multiple class-specific training sets. We refer to this case as generic fusion training. The results of generic fusion demonstrate that the proposed MMSR yields superior results to those that can be obtained by any class-specific or generic SR model on non-homogeneous test sets. Moreover, this result generalizes to arbitrary test datasets such that MMSR with pre-trained SR models (that never saw the training set for a particular test set) outperforms the best single SR model that is trained on the particular training set for the given test set.


\vspace{-8pt}
\section{Experiments}
\label{sec:experiment}
\vspace{-3pt}

In this section, the performance of class-specific SR models and the proposed MMSR framework are evaluated and compared with those of a generic EDSR model \cite{EDSR2017} and image-specific zero-shot SR models \cite{ZSSR}. 

In our experiments, we used ImagePairs \cite{joze2020imagepairs}, DIV2K \cite{Agustsson_2017_CVPR_Workshops} and RealSR-v3 \cite{caiNTIRE2019} datasets. ImagePairs is a real SR dataset obtained by using a beam-splitter to capture the same scene by a LR camera and a HR camera. RealSR dataset contains real-world scenes with sharper edges and finer textures. DIV2K is a commonly used benchmark SR dataset consisting of high-texture images and corresponding simulated LR images. We created three subsets of HR images from these three datasets, namely homogeneous text-class set, homogeneous texture-class set, and a non-homogeneous image set containing both text and texture within the same image. Corresponding LR images are obtained by $\downarrow$4 bicubic downscaling and 40 dB white Gaussian noise is added to downsampled images. 

\begin{table}[t]
\begin{center}
\caption{The mean and variance of PSNR (var in parenthesis) for the generic EDSR model on benchmark test datasets. \vspace{6pt}}
\begin{adjustbox}{width=0.45\textwidth}
\begin{tabular}{|c|l|l|l|}
\hline
Dataset  & \hspace{20pt} x2  & \hspace{20pt} x3 & \hspace{20pt} x4 \\
 \hline
 Set5 & 37.99 (8.4) & 34.37 (5.94) & 32.09 (5.53) \\
 \hline
 Set14 & 33.57 (15.62) & 30.28 (15.15) & 28.58 (14.11) \\
  \hline
 B100 & 32.15 (17.86) & 29.08 (15.95) & 27.56 (14.31) \\
 \hline
 Urban100 & 31.98 (22.66) & 28.15 (19.36) & 26.03 (16.72) \\
 \hline
 DIV2K & 34.61 (23.5) & 30.92 (23.19) & 28.95 (22.19) \\
 \hline
\end{tabular}
\end{adjustbox}
\label{table:edsr_variance}
\end{center}   \vspace{-6pt}
\end{table}

\begin{table*}[t!]
\begin{center}
\caption{Performance of generic and specific models on class-specific test datasets. Variance of PSNR is shown in parenthesis.}
\begin{adjustbox}{width=\textwidth}
\begin{tabular}{|c|c|c|c|c|c|c|c|c|}
\hline
Class  & \multicolumn{4}{|c|}{Text Class} & \multicolumn{4}{|c|}{Texture Class}\\
 \hline
Measure  & \multicolumn{1}{|c|}{PSNR} & \multicolumn{1}{|c|}{SSIM} &  \multicolumn{1}{|c|}{LPIPS-Alex} &  \multicolumn{1}{|c|}{LPIPS-VGG} & \multicolumn{1}{|c|}{PSNR} & \multicolumn{1}{|c|}{SSIM} &  \multicolumn{1}{|c|}{LPIPS-Alex} &  \multicolumn{1}{|c|}{LPIPS-VGG}
  \\ \hline
  Generic-EDSR & 32.903 (3.205) & 0.927 & 0.1171 & 0.1821 
                & 29.822 (19.266) & 0.8491 & 0.2604 & 0.2879
  \\ \hline
  Class-Specific & \textbf{33.328} (3.067) & \textbf{0.929} & 0.112 & 0.1803 & \textbf{29.829} (16.302) & \textbf{0.8493} & 0.2603 & 0.2860 
 \\ \hline 
  ZSSR \cite{ZSSR} & 24.639 (1.653) & 0.796  & \textbf{0.0829} & \textbf{0.153} 
       & 29.011 (17.828) & 0.832  & \textbf{0.118} & \textbf{0.1629}
  \\ \hline
\end{tabular}
\end{adjustbox}
\label{table:class_based_results}
\end{center}  \vspace{-18pt}
\end{table*}

\begin{table}
\caption{Performance of the MMSR model with class-specific fusion on 20-text images from the ImagePairs \cite{joze2020imagepairs}. \vspace{-18pt} }  
\begin{center}
\begin{adjustbox}{width=0.45\textwidth}
\begin{tabular}{|c|c|c|c|c|}
 \hline
 & PNSR & SSIM & LPIPS-Alex & LPIPS-VGG\\
 \hline
 Text-Specific SR & 33.328 (3.229) & 0.929 & 0.1120 & 0.1803 \\
 \hline
 Texture-Specific SR & 25.598 (2.273) &  0.838 & 0.1818 & 0.2889\\
 \hline
 Generic SR & 32.903 (3.373) & 0.927 & 0.1171  & 0.1821\\
 \hline
 Multiple-Model SR & \textbf{33.668} (3.211) & \textbf{0.932} & \textbf{0.1100} & \textbf{0.1772} \\
 \hline
\end{tabular}
\end{adjustbox}
\label{table:text_mmsr_results}
\end{center}  \vspace{-8pt}
\end{table}

During the training of all models used in this study, we selected minibatch size as 8 and each minibatch consists of randomly cropped 96$\times$96 patches from randomly selected images. We use the Adam optimizer with default parameters. The learning rate is initialized as 0.0001 and it is halved every 50k iterations. It takes almost 2-days on a Tesla V100 to train the models for 130k iterations. 

Our generic EDSR model is trained on the union of all images from all three datasets. In order to demonstrate limitations of generic SR models, we run tests for $\times$2 SR, $\times$3 SR, and $\times$4 SR tasks on 5 benchmark test datasets. Table~\ref{table:edsr_variance} shows average PSNR and variance of PSNR for a generic EDSR model \cite{EDSR2017} obtained over these test sets. It shows that the variance of PSNR values are quite large, and hence, the mean PSNR values are not very informative on the performance of the model for individual images.



\subsection{Results for Text-Class SR}
\label{result-text}

To train and test our text-class SR model, we used 80 images for training and 20 images for testing from the recently released ImagePairs \cite{joze2020imagepairs} dataset. Since all images are a snapshot from a journal, intra-class variation is small. Compared to the generic SR model, the text-specific model improves both PSNR and SSIM values as shown in Table \ref{table:class_based_results}, while also decreasing the intra-class PSNR variance. We observe that ZSSR \cite{ZSSR} has lower PSNR and SSIM values compared to the generic and text-class models, but it gives the best results in terms of the LPIPS perceptual metric \cite{zhang2018perceptual}. Even though ZSSR achieves better LPIPS scores, Fig. \ref{fig:results} shows that the output of ZSSR has significant ringing and deviates from the HR image in the vicinity of the edges of the letters. On the other hand, the text-class model, depicted in  Fig.~\ref{fig:results}(f), is able to better match edge transitions and provides superior performance.

\subsection{Results for Texture-Class SR}
\label{results-texture}

\begin{figure}[!ht]
\begin{minipage}[b]{0.72\linewidth}
  \centering
  \centerline{\includegraphics[width=6.1cm]{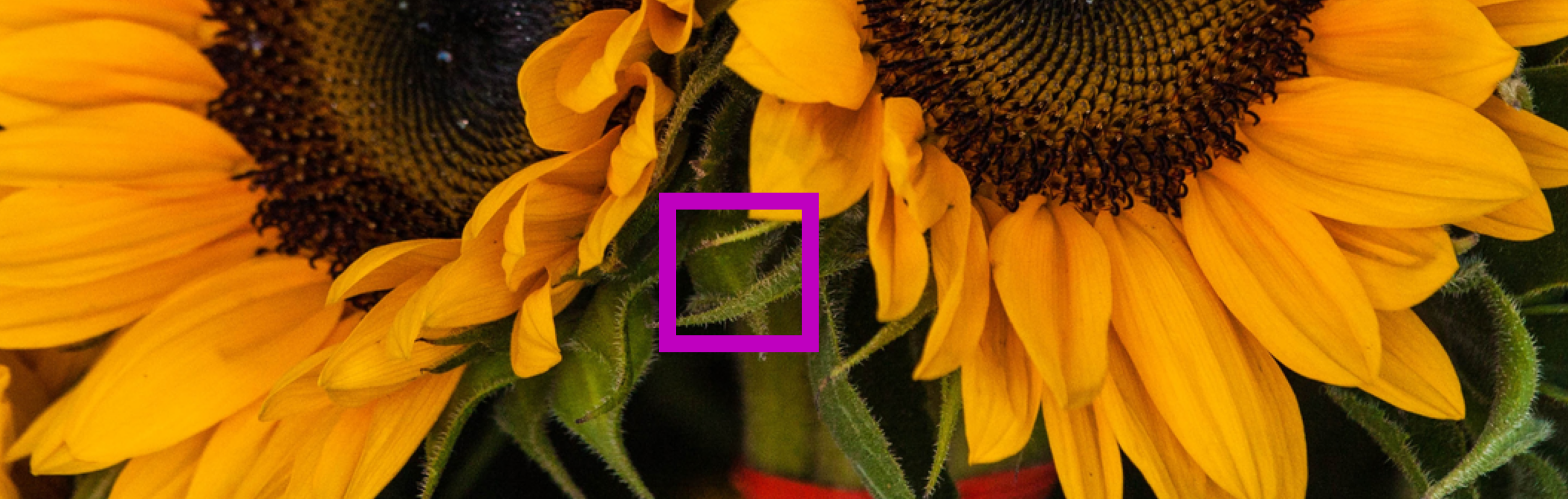}} \vspace{-3pt}
  \centerline{(a)}
\end{minipage}
\begin{minipage}[b]{.24\linewidth}
  \centering
  \centerline{\includegraphics[width=2.0cm]{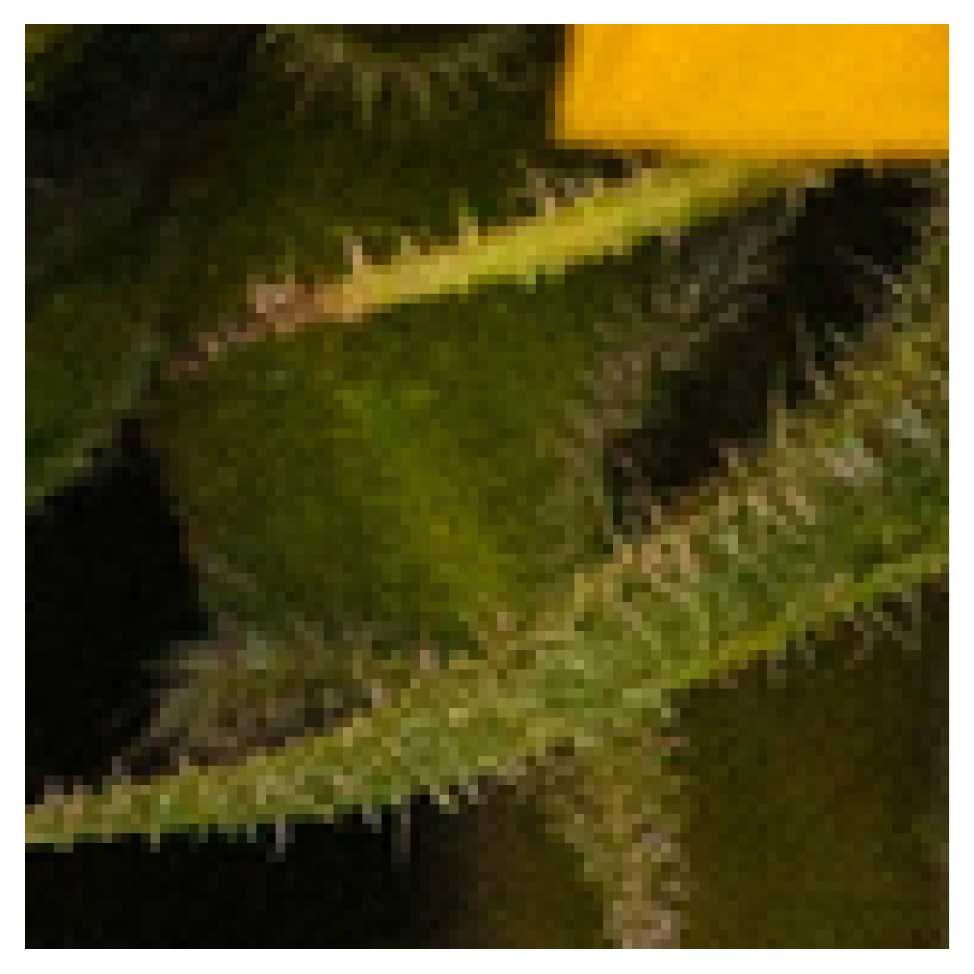}} \vspace{-3pt}
  \centerline{(b)}
\end{minipage}
\begin{minipage}[b]{.24\linewidth}
  \centering
  \centerline{\includegraphics[width=2.0cm]{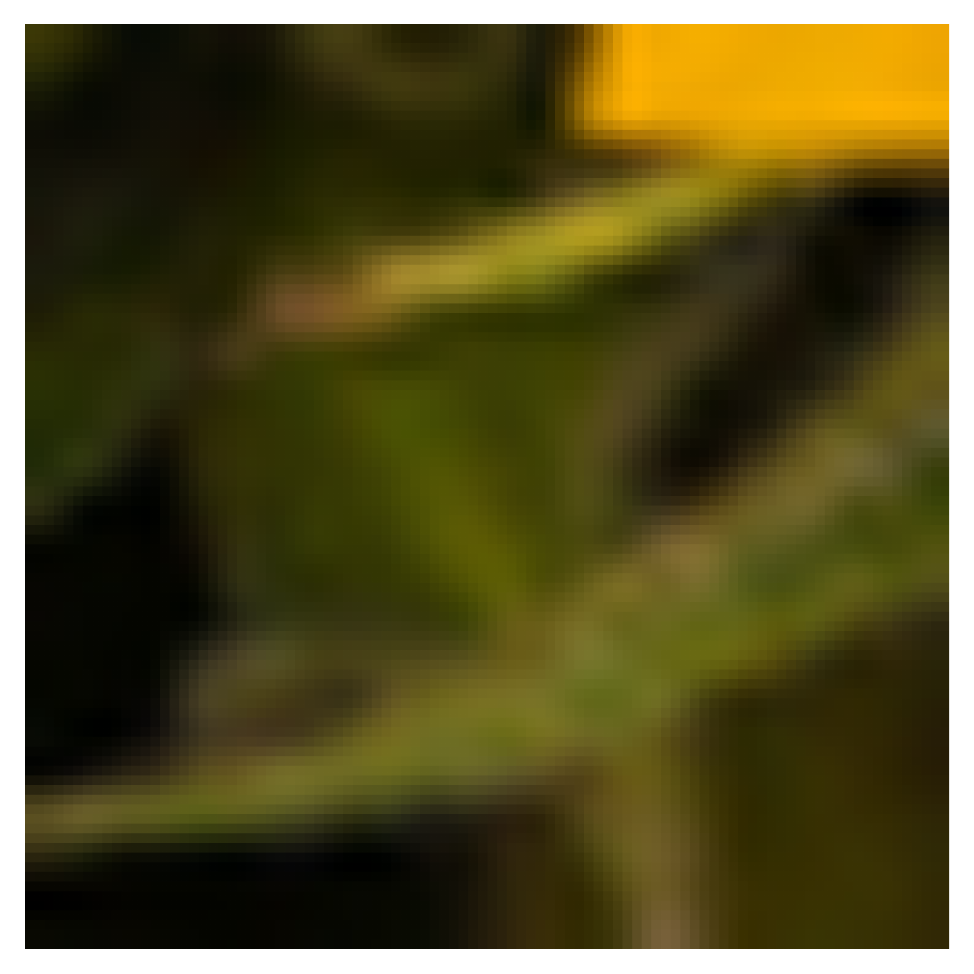}} \vspace{-3pt}
  \centerline{(c)}\medskip
\end{minipage}
\hfill
\begin{minipage}[b]{0.24\linewidth}
  \centering
  \centerline{\includegraphics[width=2.0cm]{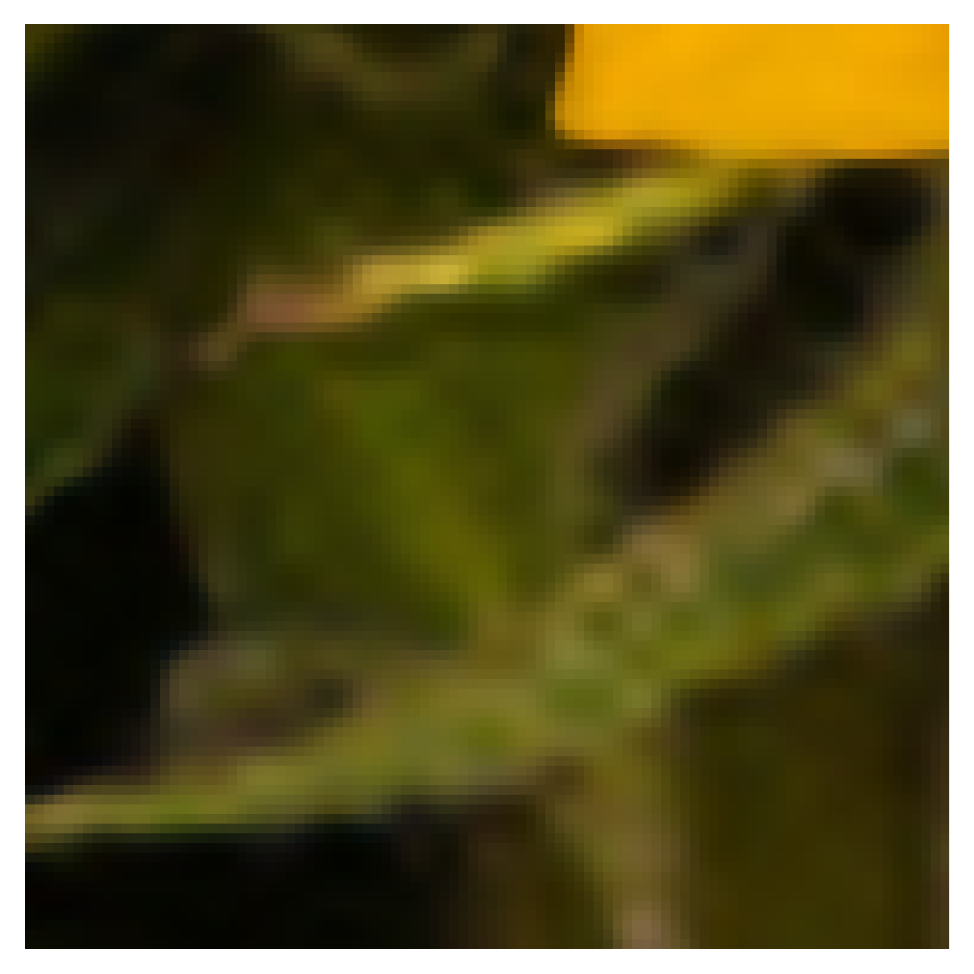}} \vspace{-3pt}
  \centerline{(d)}\medskip
\end{minipage}
\begin{minipage}[b]{.24\linewidth}
  \centering
  \centerline{\includegraphics[width=2.0cm]{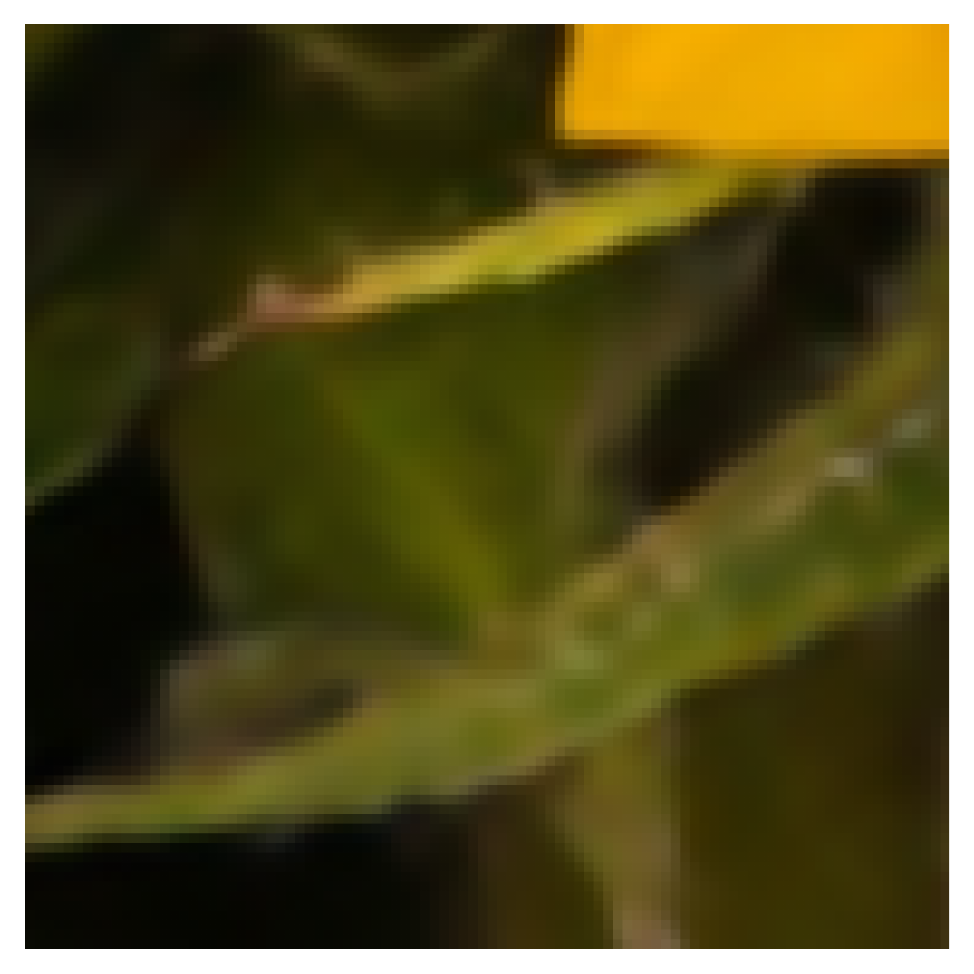}} \vspace{-3pt}
  \centerline{(e)}\medskip
\end{minipage}
\hfill
\begin{minipage}[b]{0.24\linewidth}
  \centering
  \centerline{\includegraphics[width=2.0cm]{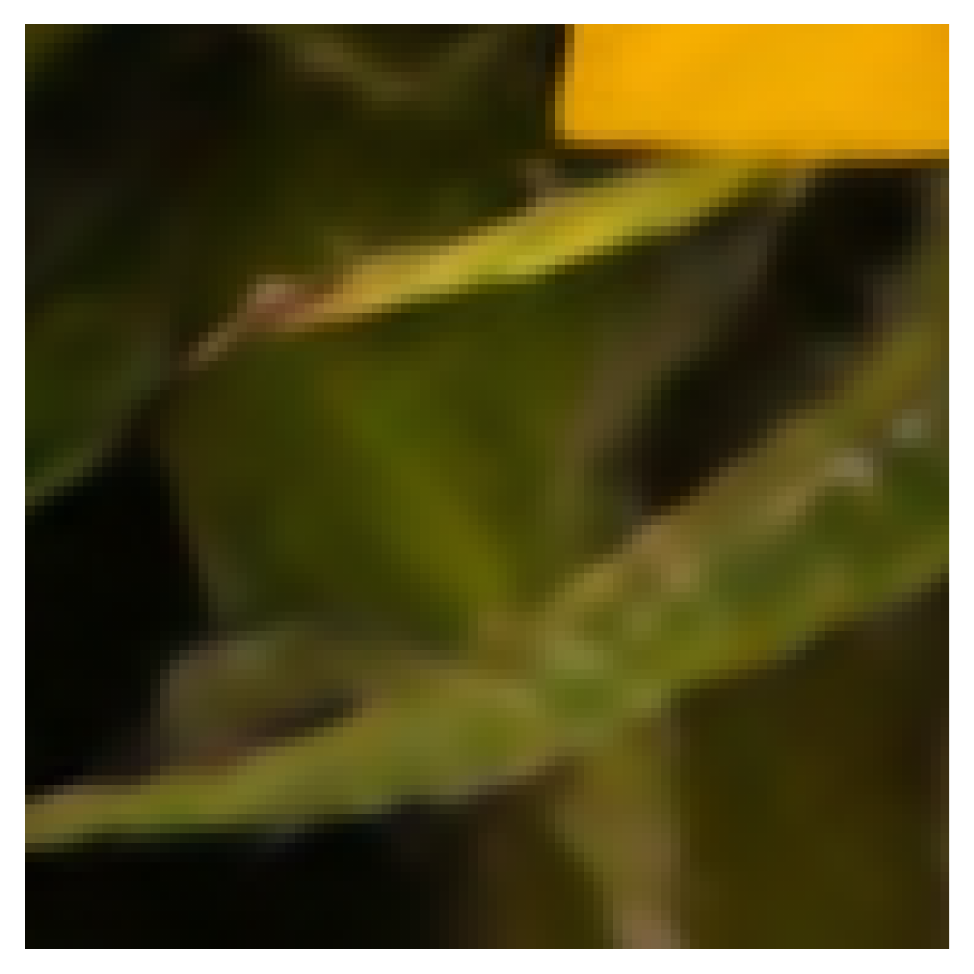}} \vspace{-3pt}
  \centerline{(f)}\medskip
\end{minipage}  \vspace{-15pt}
\caption{This example demonstrates that $\times$4 SR by texture-class model not only yields higher PSNR, but also have better visual quality. (a) HR image 0746 - DIV2K \cite{Agustsson_2017_CVPR_Workshops} test set, (b)~cropped HR patch, (c) bicubic interpolation of $\downarrow$4 image, (d)~ZSSR - 33.194 dB, (e)~generic EDSR model - 33.371 dB, (f)~texture-class EDSR model - 33.580 dB}
\label{fig:results_texture} \vspace{-4pt}
\end{figure}

We used 140 images selected from a combination of RealSR \cite{cai2019toward} and DIV2K \cite{Agustsson_2017_CVPR_Workshops} for training the texture-class model, and 35 images for testing it. We note that the texture class is less homogeneous compared to the text class, since textures can have a wide variety, including fine, medium and coarse grain textures. The results for the texture-class SR model is given in the right part of Table~\ref{table:class_based_results}, which shows that the texture-class model has less PSNR variance compared to the generic model, but it is not as low as that of the text-class model. Similar to the case of the text-class, ZSSR \cite{ZSSR} performs better in terms of the LPIPS metric. However, the texture-specific model performs better in terms of PSNR and SSIM. Results for an example image from the DIV2K validation set are visualized in Fig. \ref{fig:results_texture}. 
\vspace{-5pt}
\subsection{Results for Multiple-Model SR}
\label{ablation}
We analyze the performance of the MMSR with $N=2$ and $N=3$ models and class-specific fusion on the text-class image set selected from ImagePairs~\cite{joze2020imagepairs}.  We use the text-class, texture-class and generic models explained in Sec. \ref{sec:class_generic} as pre-trained models and train the class-specific fusion model on 80 training images from ImagePairs. Although texture-class and generic models perform poorly on text images, they collaboratively boost the performance of the MMSR network after fusion. Table~\ref{table:text_mmsr_results} validates the performance improvement of utilizing the multiple-model SR approach, which shows that MMSR benefits from various image priors, thus, surprisingly even outperforms the text-class model, which is trained exclusively on text images. 
\vspace{5pt}

\noindent \textbf{Generalization to Other Test Datasets.} The proposed MMSR network generalizes well to other test sets containing images with a mixture of text and texture characteristics. Evaluation results on 35 images from the RealSR-V3 dataset that have both text/texture and other image patterns are summarized in Table~\ref{table:mixture_mmsr_results}. It is important to note that none of the pre-trained models have seen any images from the RealSR-V3 training set. This is why all 3 models perform poorly on this test set. However, when we fuse the results of multiple SR networks tuned to class-specific image priors, the SR performance improves significantly. We can thus conclude that the MMSR approach can compensate for deficiencies of individual models, while also generalizing well to unseen types of content.

\begin{table}
\caption{Performance of MMSR model with generic fusion on 36 non-homogeneous images from the RealSR-V3 test set \cite{cai2019toward}. (EDSR and MMSR models trained on RealSR-V3 training set)} \vspace{-10pt}
\begin{center}
\begin{adjustbox}{width=0.45\textwidth}
\begin{tabular}{|c|c|c|c|c|}
\hline
 & PNSR & SSIM & LPIPS-Alex & LPIPS-VGG \\
 \hline
 Text-Specific SR & 24.415 (7.297) & 0.798 & 0.1793 & 0.2759 \\
 \hline
 Texture-Specific SR & 23.598 (8.212) & 0.775 & 0.2008 & 0.3187\\
 \hline
 Generic SR & 23.482 (6.784) & 0.803 & 0.1799 & 0.2744\\
 \hline
 Multiple-Model SR & \textbf{30.279} (7.822) & \textbf{0.909} &  \textbf{0.1032} & \textbf{0.1923}\\
 \hline
\end{tabular}
\end{adjustbox}
\label{table:mixture_mmsr_results}
\end{center}  \vspace{-8pt}
\end{table}

\vspace{-8pt}
\section{Conclusion}
\label{sec:conclude}
\vspace{-6pt}
Many different supervised deep-learning architectures have been proposed for SISR. However, they all employ a single model to super-resolve a wide variety of images. The capability of a single model to capture image priors for different types of content is limited. This results in high PSNR variance over the test set, preventing the mean PSNR from being an informative metric on the SR performance on individual test images. In this work, we first employ class-specific models to improve SR performance and reduce intra-class PSNR variance. We then propose a multiple model SR (MMSR) framework, where a post-processing network learns how to fuse the output images produced by multiple SR models each tuned to a specific class of image priors. Both qualitative and quantitative results show that the proposed MMSR framework yields not only quantitatively and qualitatively better results but also decreases PSNR variance of SR performance over test sets containing a diverse variety of image content.


\clearpage

\bibliographystyle{IEEEbib}
\bibliography{source}

\end{document}